\documentclass[sigconf,authordraft,review=false]{acmart}

\AtBeginDocument{%
  \providecommand\BibTeX{{%
    \normalfont B\kern-0.5em{\scshape i\kern-0.25em b}\kern-0.8em\TeX}}}

\usepackage{multirow}
\usepackage{graphicx,subcaption}
\usepackage{todonotes}

\def\legalbert{LegalBERT}
\def\caselawbert{CaseLawBERT}
\def\inlegalbert{InLegalBERT}
\def\incaselawbert{InCaseLawBERT}
\def\custombert{CustomInLawBERT}

\begin{document}

\title{Pre-trained Language Models for the Legal Domain: A Case Study on Indian Law}

\author{Shounak Paul}
\affiliation{%
   \institution{Indian Institute of Technology Kharagpur}
   \city{Kharagpur}
   \state{West Bengal}
   \country{India}
   \postcode{721302}
}

\author{Arpan Mandal}
\affiliation{%
   \institution{Indian Institute of Technology Kharagpur}
   \city{Kharagpur}
   \state{West Bengal}
   \country{India}
   \postcode{721302}
}

\author{Pawan Goyal}
\affiliation{%
   \institution{Indian Institute of Technology Kharagpur}
   \city{Kharagpur}
   \state{West Bengal}
   \country{India}
   \postcode{721302}
}

\author{Saptarshi Ghosh}
\affiliation{%
   \institution{Indian Institute of Technology Kharagpur}
   \city{Kharagpur}
   \state{West Bengal}
   \country{India}
   \postcode{721302}
}









\newcommand{\bh}[1]{{\leavevmode\color{blue}{#1}}}
\newcommand{\red}[1]{{\leavevmode\color{red}{#1}}}

\begin{abstract}
NLP in the legal domain has seen increasing success with the emergence of Transformer-based Pre-trained Language Models (PLMs) pre-trained on legal text. 
PLMs trained over European and US legal text are available publicly; however, legal text from other domains (countries), such as India, have a lot of distinguishing characteristics. 
With the rapidly increasing volume of Legal NLP applications in various countries, it has become necessary to pre-train such LMs over legal text of other countries as well. 
In this work, we attempt to investigate pre-training in the Indian legal domain. 
We re-train (continue pre-training) two popular legal PLMs, LegalBERT and CaseLawBERT, on Indian legal data, as well as train a model from scratch with a vocabulary based on Indian legal text. We apply these PLMs over three benchmark legal NLP tasks -- Legal Statute Identification from facts, Semantic Segmentation of Court Judgment Documents, and Court Appeal Judgment Prediction -- over both Indian and non-Indian (EU, UK) datasets. We observe that our approach not only enhances performance on the new domain (Indian texts) but also over the original domain (European and UK texts). We also conduct explainability experiments for a qualitative comparison of all these different PLMs.
\end{abstract}





\maketitle

\section{Introduction} \label{sec:intro}


The introduction of Transformer-based Pre-trained Language Models (PLMs) and architectures such as BERT~\citep{devlin2019bert} has accelerated research in Natural Language Processing (NLP). 
In fact, pre-training language models on large corpora (which can be thought to be like large swathes of unlabeled text) has shown to be effective in many downstream tasks.
Although earlier efforts mostly utilized general (open) domain data for pre-training~\citep{liu2019roberta,yang2019xlnet,beltagy2020longformer}, limited ability of such PLMs to fine-tune to end tasks on specialized domains such as biomedical or law has prompted efforts in domain-specific pre-training~\citep{beltagy2019scibert,lee2019biobert}.

Particularly, NLP is increasingly being applied in the legal domain, and naturally there have been efforts to develop PLMs for the legal domain. \citet{chalkidis2020legalbert} pre-trained \textit{BERT-base} on legal documents from EU, UK and US. This model has been shown to be effective in many downstream legal tasks, such as text classification, summarization and named entity recognition, among others.
Again, \citet{zheng2021legalbertus} pre-trained \textit{BERT-base} on only US case law documents, which was shown to improve performance over US legal datasets such as CaseHOLD. 
More recently, \citet{henderson2022pileoflaw} pre-trained \textit{BERT-large} on `Pile of Law' (PoL), a huge legal corpus of US and EU documents.\footnote{Although all these three models were named as LegalBERT in the original research papers, we shall address them as LegalBERT~\citep{chalkidis2020legalbert}, CaseLawBERT~\citep{zheng2021legalbertus} and PoLBERT~\citep{henderson2022pileoflaw} respectively, for sake of comprehension.}
Finally, the Lawformer model~\citep{xiao2021lawformer}, pre-trained on Chinese legal text, has been designed keeping in mind that legal documents are usually much longer than documents in the general domain. 

With the advent of digitization and image processing technologies, a lot of legal data has become readily available in digital format across many countries, like India.
This has led to several new studies and technologies being developed on Indian legal data in recent years~\citep{bhattacharya2019rhetoricalrole,bhattacharya2019comparative,malik2021ildc,paul2022lesicin,kalamkar2021indianlegalnlp}.
Using BERT off-the-shelf for these tasks has been shown to be limited in performance, and although LegalBERT improves upon BERT across many such tasks, there is still scope for improvement in tasks of Indian legal text.
This has to do particularly with the  differences in the nature of Indian legal texts vis-a-vis EU and US legal texts~\citep{kalamkar2021indianlegalnlp,bhattacharya2019comparative}. 
Also, due to typographical or scanning errors, there are a lot of instances of missing spaces and punctuation marks, which lead to wrong words.
Thus, it is beneficial to pre-train LMs over Indian legal text, allowing the model to observe such patterns during pre-training, so that performance over end-tasks over Indian legal text improves.

In this work, we conduct pre-training experiments with PLMs on a large Indian legal corpus.
We collected a large corpus of court judgment documents from the Supreme Court and many High Courts of India, along with a collection of all Central Government Acts (statutory documents).
Our dataset is decently positioned in terms of the number of documents (\textasciitilde 5.4M) and size (\textasciitilde 27GB) as compared to other pre-training datasets used in the legal domain (see Table~\ref{tab:ptdatasets}).
We pre-train BERT-based LMs on this corpus and estimate the efficacy of domain-specific pre-training on a number of downstream end-tasks on both Indian legal documents as well as legal documents from other countries/domains.

\begin{table*}[tb]
    \small
    \centering
    \begin{tabular}{|p{0.2\textwidth}|p{0.5\textwidth}|c|c|}
        \hline
        \textbf{Model} & \textbf{Data Source} & \textbf{\#Documents} & \textbf{Size (GB)} \\ \hline
         LegalBERT & EU, UK, US legislation, cases from ECJ, ECtHR  & 350K & 12  \\ \hline
         CaseLawBERT & Harvard Case Law (based on US federal and state courts) & 3.4M & 37 \\ \hline
         PoLBERT & Legal Analyses, Court Opinions, Government Publications, Contracts, Statutes, Regulations, and more from US and EU & 10M & 256 \\ \hline
         Lawformer & China Judgment Online (based on cases from Chinese Courts) & 22.7M & 84 \\ \hline \hline
         \inlegalbert{}, \incaselawbert{}, \custombert{} & Cases from Indian Supreme Court and High Courts & 5.4M & 27 \\ \hline
    \end{tabular}
    \caption{Comparison of pre-training datasets used in various works in the legal domain. The last row is about the Pre-trained Language Models (PLMs) developed in this work.}
    \label{tab:ptdatasets}
    \vspace{-5mm}
\end{table*}

\noindent {\bf Our Contributions:} To enumerate briefly, our contributions are as follows:

\noindent(i)~We re-trained (continued pre-training) \footnote{We use the terms `re-train' and `continue pre-training' or `further pre-training' interchangeably throughout the text.} two publicly available pre-trained BERT-based legal PLMs, namely LegalBERT~\citep{chalkidis2020legalbert} and CaseLawBERT~\citep{zheng2021legalbertus} on a large corpus of Indian legal corpus.
Both the Indian PLM variants -- which we name \inlegalbert{} and \incaselawbert{} respectively -- outperform their source LMs in terms of perplexity on Indian data.

\noindent (ii)~We also pre-trained a BERT-base model from scratch on this Indian legal pre-training corpus, with a custom vocabulary specifically tailored for Indian legal text. We name this model \custombert{}.

\noindent(iii)~We compare these three Indian legal PLMs among themselves, as well the original LegalBERT and CaseLawBERT, on several practically important end-tasks, namely, (1)~Legal Statute Identification~\citep{paul2022lesicin,chalkidis2022lexglue}, (2)~Semantic Segmentation of judgements~\citep{bhattacharya2019rhetoricalrole,bhattacharya2021deeprhole}, and (3)~Court Judgement Prediction~\citep{malik2021ildc}. 
We perform experiments on different benchmark datasets (based on both Indian and non-Indian legal texts) for these tasks. 
Our experiments broadly show that \inlegalbert{} achieves appreciable gains over LegalBERT across all tasks (while the gains are minimal for \incaselawbert{} over CaseLawBERT).
We also observe that \custombert{} comes close to the performance of both \inlegalbert{} and \incaselawbert{} on multiple datasets, despite having been pre-trained for a much lesser number of steps (700k vs. 1.3M vs. 2.3M steps respectively).

\noindent (iv)~To understand the inherent differences between the various PLMs, we analyze the model attentions for the Court Judgement Prediction task with a small set of expert-annotated documents (made available in~\citep{malik2021ildc}). 
Comparing model-assigned attention scores with expert-annotated scores shows that the correlation between these scores increases with pre-training on Indian data.

Out of all the pre-trained Language Models (PLMs) that we experiment with, the \inlegalbert{} (developed in this work) outperforms all other PLMs in almost all the end-tasks that we experimented with.
Additionally, comparing with the original works which contributed the respective datasets, \inlegalbert{} establishes state-of-the-art results on 4 out of 5 datasets.
These results show that further pre-training on Indian legal data helps in improving performance over English-language legal text datasets from not only India but from other countries as well. 


All the three models are publicly available on HuggingFace.\footnote{
\inlegalbert{} --- \url{https://huggingface.co/law-ai/InLegalBERT} \\
\incaselawbert{} --- \url{https://huggingface.co/law-ai/InCaseLawBERT} \\
\custombert{} --- \url{https://huggingface.co/law-ai/CustomInLawBERT}
}
The code for pre-training are also available on GitHub. \footnote{\url{https://github.com/Law-AI/pretraining-bert}}


\section{Related Work} \label{sec:relwork}
\noindent \textbf{Transformer-based PLMs:} 
After the introduction of the self-attention based Transformer mechanism~\citep{vaswani2017transformers}, we have seen the widespread development of large-scale, pre-trained language models.
\citet{devlin2019bert} first introduced BERT, a multi-layer transformer architecture for language modeling. 
They used the Masked Language Modeling (MLM) objective \citep{taylor1953cloze} to mask some tokens randomly in the text, to effectively train a deep bidirectional language model on general domain corpora, including English Wikipedia.
They also used Next Sentence Prediction (NSP) objective to sample pairs of contiguous sentences.
Their proposed approach of transfer learning differed from prior works in the sense that most earlier works used to transfer pre-trained embeddings for the end tasks, whereas BERT proposed transferring the pre-trained model instead.

The unprecedented success of BERT on a range of downstream NLP tasks such as \texttt{GLUE} \citep{wang2018glue} and \texttt{SQUAD} \citep{rajpurkar2016squad} prompted research in transformer based PLMs.
RoBERTa \citep{liu2019roberta} used a much larger pre-training corpus and did away with the NSP objective, going for larger batch sizes instead. 
XLNet \citep{yang2019xlnet} modified the pre-training objective to Permutation Language Modeling.
Longformer \citep{beltagy2020longformer} introduced a global-local attention mechanism to deal with longer sequences.
All these models were shown to outperform BERT on many general domain downstream tasks.
However, these PLMs, pre-trained on general domain data, often show limited performance for tasks in specialized domains such as biomedical and law, leading to efforts in pre-training such models on domain-specific data, e.g., BioBERT~\citep{lee2019biobert} and SciBERT~\citep{beltagy2019scibert}.


\vspace{2mm}
\noindent \textbf{PLMs for the legal domain:}
In recent years, a lot of legal tasks and datasets have been released for multiple domains/countries, such as US, EU, China, India and so on. 
This has led to use of pre-trained transformer based models such as BERT \citep{devlin2019bert} on these datasets.
Legal text being very unique, 
there is significant scope of improvement over models such as BERT pre-trained on the general domain. 
There have been multiple efforts to pre-train transformers on the legal domain:
(i)~\citet{chalkidis2020legalbert} pre-trained BERT-base on EU and UK legislation and court documents from the US, European Court of Justice (ECJ) and European Court of Human Rights (ECtHR), and released the LegalBERT model;
(ii)~\citet{zheng2021legalbertus} proposed CaseLawBERT, pre-trained on a corpus of US case law documents and contracts;
(iii)~\citet{henderson2022pileoflaw} prepared a huge corpus of US, Canada and EU documents (not just case law), called \emph{Pile of Law}, and trained BERT-large on the same to yield the PoLBERT model;
(iv)~\citet{xiao2021lawformer} released Lawformer, a Longformer~\citep{beltagy2020longformer} based model pre-trained on Chinese legal text.
The details of the pre-training datasets used by these models are available in Table~\ref{tab:ptdatasets}.

However, no work has compared the performances of these different PLMs in a common setting. In addition, it has not been explored how these existing PLMs can be utilized to develop PLMs for legal text of other domains/countries. We attempt to fill these gaps in this paper, with a case-study on Indian legal text.

\section{Data Collection} \label{sec:data}

For pre-training, we collected court case documents from different courts of India, specifically, the Supreme Court, 26 different High Courts and 2 different District Courts (total of 29 courts).
We used publicly available sources such as official websites of these courts (e.g., the website of the Indian Supreme Court: \url{https://main.sci.gov.in/}), the erstwhile website of the Legal Information Institute of India, 
the popular legal repository \url{https://www.indiankanoon.org}, and so on.
Additionally, we collected 1,113 Central Government Acts, which are the documents codifying the laws of the country. Each Act is a collection of related laws, called Sections. These 1,113 Acts contain a total of 32,021 Sections. 

In total, we collected around \emph{5.4 million documents}, comprising of around \emph{27GB raw text}.
A brief comparison of our pre-training dataset with those used in prior works in given in Table~\ref{tab:ptdatasets}.

The court cases in our dataset range primarily from 1950 to 2019 (India became a republic in 1950), although there are a few cases dating as far back as the 1800s.
These cases belong to all legal domains, such as Civil, Criminal, Constitutional, and so on.
The temporal distribution of cases in our dataset is heavily skewed towards the recent decades.
This skew is natural, since cases have risen exponentially with India's growing population, as well as the increasing application of digital techniques in the judicial setup for keeping records.

\vspace{2mm}
\noindent {\bf Pre-Processing the documents:}
As discussed in Section~\ref{sec:intro}, Indian legal text is known to be noisy.
There are many odd artefacts in the text, such as sequences of dashes or lines, page numbers and metadata in random positions of the text, etc.
Most of such anomalies have crept into the text due to digitization of hand-written documents (which we verified manually by comparing older documents with newer ones, especially post 2010).
We remove such artefacts during pre-training using Regular Expression matching.

\section{Pre-training Setup} \label{sec:pretrain}

We now describe our approach for developing a PLM for Indian legal text.
We conduct different types of independent pre-training experiments ---

\noindent (i)~by starting with the two existing pre-trained models -- LegalBERT~\citep{chalkidis2020legalbert} \footnote{\url{https://huggingface.co/nlpaueb/legal-bert-base-uncased}} and CaseLawBERT~\citep{zheng2021legalbertus} \footnote{\url{https://huggingface.co/zlucia/legalbert}} -- 
and continuing the pre-training on our Indian dataset (Re-training).
We name the two final models (after re-training on the Indian dataset) as \inlegalbert{} and \incaselawbert{}, respectively. 

\noindent (ii)~by training a BERT-base-uncased model from scratch, and utilizing a custom vocabulary trained on our data. 
We name this model as \custombert{}.

\begin{figure}[t]
\centering
\includegraphics[width=0.6\columnwidth,height=4cm]{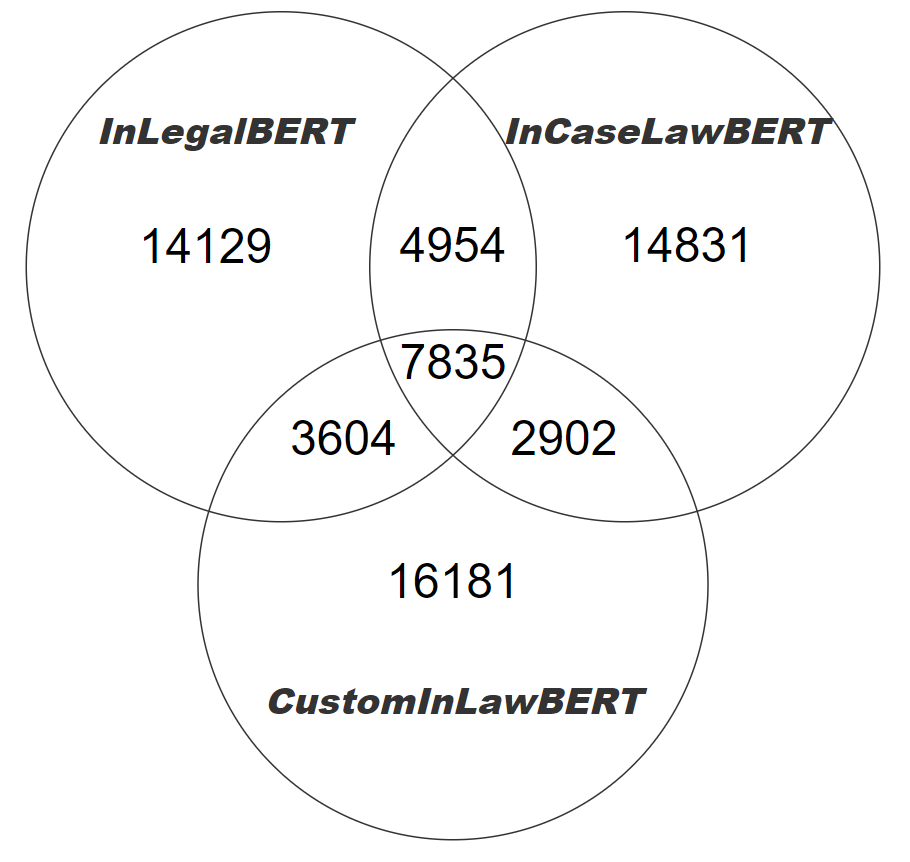} 
\caption{Comparison of vocabularies of the three PLMs -- \inlegalbert{} (vocabulary same as that of LegalBERT), \incaselawbert{} (vocabulary same as that of CaseLawBERT) and our custom vocabulary for Indian legal text.}
\label{fig:vocab}
\vspace{-5mm}
\end{figure}

\subsection{Custom Vocabulary for Indian Legal Data}

The language of Indian legal documents is quite different from legal documents of other countries, even though all are written in English.
One of the major reasons for this difference is the frequent use of local terms like ``daroga'' (constable) or ``lathi'' (stick), or local salutations like ``Shri'' (Mr.) or ``Smt.'' (Mrs.), and so on. 
These factors motivate the need of using a custom vocabulary tailored for Indian legal documents, which we follow for the experiment that involves pre-training from scratch (for developing \custombert{}).

We use the WordPiece algorithm~\citep{wu2016wordpiece} employed by BERT for training our custom tokenizer. We choose 10\% of the training data to train this tokenizer. 
We keep all the default parameters used for training the BERT tokenizer (including the vocabulary limit of 30,522 tokens).

Figure~\ref{fig:vocab} shows a quantitative comparison between the vocabularies of \inlegalbert{} (same as that of LegalBERT), \incaselawbert{} (same as that of CaseLawBERT) and our custom vocabulary.
There is a significant number of tokens that are common to all three vocabularies, which represent common English words.
It is also interesting to note that the number of tokens common to two vocabularies but not to the third is significantly lower than the tokens common to all.
On the other hand, each of the three vocabularies has a substantial number of unique tokens which are {\it not} contained in the other vocabularies. 
Also note that there is a slightly higher degree of overlap between the vocabularies of InLegalBERT and InCaseLawBERT, as compared to their overlap with our custom Indian vocabulary. This highlights the presence of unique words in Indian legal data.

\subsection{Creating training and testing examples for pre-training}
We split the corpus into train and test splits.
The test split is used to evaluate the pre-training quality using the perplexity metric (details later in this section).
The case documents were distributed randomly into the two splits in the ratio of 9:1, while maintaining the distribution of document lengths (in terms of no. of sentences) across both splits.

To create individual training and testing examples, we first divided each document $D$ into multiple chunks $\{C_1, C_2, \ldots, \}$.
For documents in both train and test dataset, we divided each document into a set of equal-sized disjoint chunks of contiguous text, disregarding sentence boundaries. 
Each chunk is a unique training/testing example. 
This strategy helps us train on longer sequences, while reducing training time~\citep{devlin2019bert}. 
Chunking also does not affect the two pre-training objectives --- MLM and NSP.
Following this strategy, we end up with a total of \emph{21.6M training examples}, and \emph{2.4M testing examples}.

\subsection{Training Objectives}

We use the standard Masked Language Modeling (MLM) and Next Sentence Prediction (NSP) training objectives that were also used in pretraining BERT~\cite{devlin2019bert}.
MLM involves masking some tokens of each input during encoding, and the model should predict the masked tokens based on the context.
Under NSP, a pair of inputs is passed to BERT, and the model is required to predict whether the second input follows the first in the document sequence. 
We use both \textit{dynamic masking} for MLM \citep{liu2019roberta} and \textit{dynamic sampling} for NSP, meaning that the choice of masked words (MLM) and negative samples (NSP) are \textit{not} fixed beforehand in a static way; rather, it is dynamically chosen when a mini-batch is being built. 
For each training example $C_i \in D$ (which is a chunk of a document, as stated above), we either pair it up with its consecutive chunk $C_{i+1}$ (positive sample) or randomly choose any other chunk from the same document $C_j, j \neq i, i+1$ (negative sample) in a 1:1 ratio. 
This gives us the sequence pair for the NSP task.

Then, for each such pair of chunks $(C_i, C_j)$, we mask some tokens by following the masking policy described by \citet{devlin2019bert}.
Specifically, for each input pair, we randomly choose 15\% tokens across both inputs as mask candidates. 
Models like BERT use a tokenization scheme which can split words into sub-word units. We use whole-word masking, i.e., ensuring that candidate tokens are chosen in a manner where all sub-word tokens of a real word are chosen, or none at all.
Out of all such candidate tokens, we replace them with the \texttt{[MASK]} token 80\% of the time, keep them unchanged 10\% of the time, and replace them with a random token in the vocabulary 10\% of the time.

The masked sequence pair $(C'_i, C'_j)$ is passed to the BERT model along with the special tokens \texttt{[CLS]} and \texttt{[SEP]}, yielding a hidden embedding for each token $t_k$ in the entire sequence pair.
\[Input \leftarrow \texttt{[CLS]} \ C_{i1} \ C_{i2} \ldots C_{i|C_i|} \ \texttt{[SEP]} \ C_{j1} \ C_{j2} \ldots C_{j|C_i|} \ \texttt{[SEP]}\]
\[\mathbf h_k \leftarrow BERT\left( t_k \right) \ \forall t_k \in Input\]
From the BERT outputs, we pass the hidden embeddings of every token through a linear classifier for MLM.
\[P\left(\hat{y_k}^{(MLM)} = v\right) \leftarrow softmax  \left( ff_{MLM} (\mathbf h_k)  \right) \ \forall v \in V\]
where $V$ represents the vocabulary, $ff$ represents a linear layer.

For NSP, we use the BERT pooler output, which is basically the output embedding of the \texttt{[CLS]} token further passed through a feedforward network with {\sl tanh} activation. 
The pooler output is passed through another linear classifier for NSP.
\[P\left(\hat{y}^{(NSP)} = c\right) \leftarrow softmax \left(ff_{NSP} \left( tanh \left( ff_{pooler} (\mathbf h_{[CLS]}) \right) \right) \right)\]
where $c \in \{0,1\}$ represents whether the second input follows the first or not.
Both MLM and NSP are trained using the Categorical Cross Entropy Loss. 
For MLM, the loss is calculated \emph{only} over the masked tokens.

\subsection{Implementation Details}
Both the existing PLMs (used for re-training) as well as our custom PLM (trained from scratch) are based on the BERT-base-uncased model (12 layers, 768 hidden dimensionality, 12 attention heads, 110M parameters).
Due to the constraints on input lengths of BERT-based models, all chunks are limited to 512 tokens (including special \texttt{[CLS]} and \texttt{[SEP]} tokens). 
To protect most information from truncation, we chunk training documents into segments of 254 tokens each.
Thus, while sampling pairs for NSP, we ensure the total length of the sequence pair is limited to 512.
All experiments are performed on a single Nvidia RTX A6000 (48 GB VRAM).

For re-training, we train for 300K optimization steps, which is roughly equivalent to 4 full epochs over the training split.
The model being trained from scratch naturally requires more training steps to make it comparable to the re-training setups, and thus we train this model for 700K optimization steps.
It is important to note that our setup is still quite limited compared to prior works, which employed a larger number of pre-training steps for both re-training as well as pre-training from scratch.
For instance, \citet{chalkidis2020legalbert} performed 500k pre-training steps for re-training BERT, and 1M steps for a model trained from scratch.
\citet{zheng2021legalbertus} performed 1M steps when re-training BERT and 2M steps for pre-training from scratch.
At the end of our pre-training effort, we end up with models that have been trained for vastly different total no. of steps:
\inlegalbert{} (total 1.3M steps), \incaselawbert{} (total 2.3M steps), \custombert{} (total 700k steps).

We used the default AdamW optimizer~\citep{loshchilov2019adamw} with initial learning rate of 5e-5.
Apart from these, we kept all other hyper-parameters to their default values (i.e., values used by \citet{devlin2019bert} for pre-training BERT).
We also used other techniques to speed up training, such as fp16 training, pinning the CPU memory for GPU transfer, and using 8 CPU workers for preparing the batches.
In our setup, each re-training experiment took approximately 20 days (300K steps), while training from scratch took approx. 50 days (700K steps). 
Testing using perplexity took approx. 1 day.
We have had to use limited no. of pre-training steps (as discussed above) mainly because it takes such a lot of time to perform one pre-training experiment on a modest computational setup as ours.

For pre-training (and eventually fine-tuning) tasks, we used the \texttt{datasets} package \citep{lhoest2021hfdatasets} by HuggingFace (HF) for dealing with loading, tokenizing and formatting. 
We also used the \texttt{transformers} package \citep{wolf2020hftransformers} by HF to load the models and run training and evaluation tasks.
We used our own code since the default BERT pre-training code provided by \citet{devlin2019bert} do not perform dynamic MLM and NSP.
Our entire code base is built on PyTorch~\citep{paszke2017pytorch}.



\begin{figure*}[t]
\centering
\includegraphics[width=0.65\textwidth,height=5.5cm]{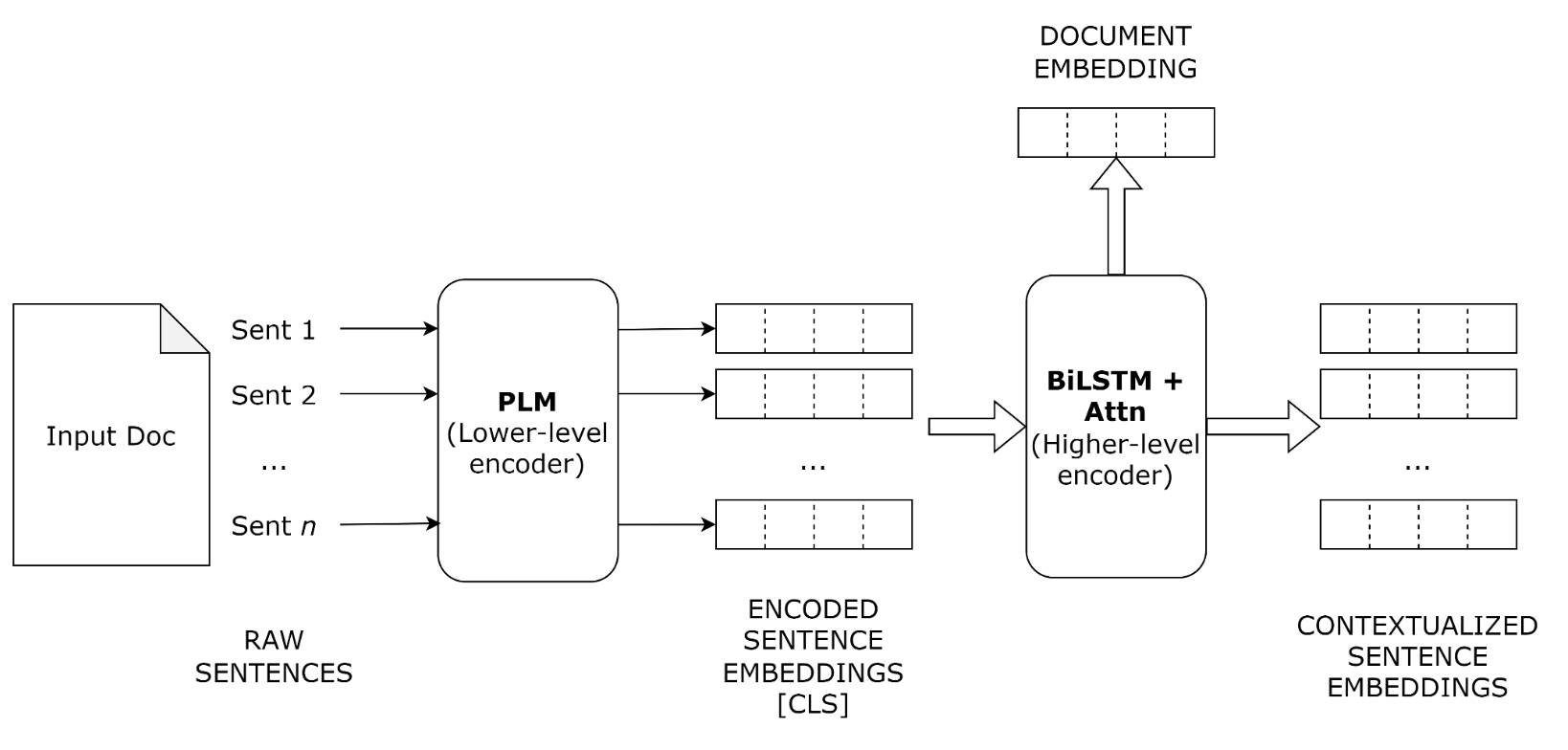} 
\vspace{-2mm}
\caption{The HierBERT architecture that is used for all end-tasks. We experiment with different versions of this architecture which use different PLMs (e.g., LegalBERT, \inlegalbert{}, \incaselawbert{}, \custombert{}, etc.) for the lower-level encoder. The rest of the architecture is kept the same in all versions, for a fair comparison among the different PLMs.}
\label{fig:ftarch}
\end{figure*}

\subsection{Evaluation of pre-training} 

To get an idea of the quality of pre-training, we evaluate the pre-trained models on the test set (10\% of the corpus, as stated earlier). 
We evaluate the language modeling capabilities of the models using the \textit{perplexity} metric.
As mentioned previously, LMs, in general, learn to estimate the probability of a sequence of tokens. 
The perplexity metric scores the model based on these probabilities.
For models that use MLM, the models can be scored based on the probabilities assigned to the masked tokens.
Mathematically, perplexity can be calculated as the exponentiation of the average Negative Log-Likelihood (NLL).
Thus, a lower value of perplexity indicates a better score for the LM.
In our case, we can directly use the loss \emph{only} for the MLM task ($\mathcal L_\mathrm{MLM}$) ---
$Perplexity = \mathrm{e}^{\mathcal L_\mathrm{MLM}}$

Table~\ref{tab:perp} shows the perplexity values of the various models over our test set.
We observe lower perplexities for the Indian variants \inlegalbert{} and \incaselawbert{}, as compared to the original PLMs.
In particular, \inlegalbert{} seems more adaptable to the Indian scenario (lower perplexity) as compared to \incaselawbert{}.
\custombert{} does not achieve as low perplexities as the other legal PLMs, but it still does better than BERT.

\begin{table}[tb]
    \centering
    \small
    \begin{tabular}{|l|r|}
        \hline
         \textbf{Model} & \textbf{Perplexity} \\ \hline
         BERT & 25.7600 \\
         LegalBERT & 7.1331 \\
         CaseLawBERT & 11.1798 \\\hline
         \inlegalbert & 5.2547 \\ 
         \incaselawbert & 8.7824 \\ 
         \custombert{} & 11.7549 \\ \hline
    \end{tabular}
    \caption{Quality of pre-training measured via perplexity. All perplexity values reported over our test set (10\% of the Indian legal text corpus).}
    \label{tab:perp}
    \vspace{-10mm}
\end{table}

\section{Application on end-tasks} \label{sec:finetune}
We now apply (fine-tune) the pre-trained models on several practically important and benchmark end-tasks over both Indian legal text as well as legal text from other domains/countries.
While these end-tasks may eventually require more complex architectures, all such architectures contain an inherent BERT-based encoder module for encoding the text either in parts or full. We try out the different PLMs, used in this study, as this encoder module, for a fair comparison.

We chose the standard BERT-base~\citep{devlin2019bert}, the original LegalBERT~\citep{chalkidis2020legalbert} and the original CaseLawBERT~\citep{zheng2021legalbertus} as baselines.
The latter two models are based on the same architecture as BERT-base.\footnote{
For the sake of fair comparison, we did \textit{not} choose PoLBERT~\citep{henderson2022pileoflaw} as a baseline since it is based on BERT-large, which is inherently more powerful.}

\begin{table*}[!t]
    \centering
    \small
    \begin{tabular}{|l||c|c|c||c|c|c|}
        \hline
        \multirow{2}{*}{\textbf{Encoder Module}} & \multicolumn{3}{c||}{\textbf{\texttt{ILSI} dataset}} & \multicolumn{3}{c|}{\textbf{\texttt{ECtHR-B} dataset}} \\ \cline{2-7}
        & \textbf{mP} & \textbf{mR} & \textbf{mF1} & \textbf{mP} & \textbf{mR} & \textbf{mF1} \\\hline
        BERT & 82.12 & 49.07 & 59.11 & 77.50 & 69.31 & 72.95 \\
        LegalBERT & \textbf{83.98} & 53.83 & 63.89 & 80.85 & 70.76 & 75.09 \\
        CaseLawBERT & 82.89 & 54.72 & 64.53 & 82.37 & 66.45 & 72.87 \\ \hline
        \inlegalbert & 82.42 & 55.16 & \textbf{64.58} & \textbf{83.93} & \textbf{71.41} & \textbf{75.88} \\
        \incaselawbert & 81.07 & \textbf{55.64} & 64.44 & 77.35 & 69.45 & 72.86 \\ 
        \custombert & 82.48 & 54.46 & 64.29 & 79.64 & 67.09 & 72.56 \\ \hline
        Best result in original paper & 27.90 & 31.32 & 28.45 & - & - & 74.70 \\ \hline
    \end{tabular}
    \caption{Performance over the \texttt{ILSI} and \texttt{ECtHR-B} datasets for Legal Statute Identification task. All reported values are macro-averaged and in terms of percentage. Difference in m-F1 results for \inlegalbert{} and all other variants is statistically significant based on the Wilcoxon Signed Rank Test.}
    \label{tab:lsires}
    \vspace{-5mm}
\end{table*}

\subsection{The fine-tuning architecture used for all end-tasks} \label{sec:ft/arch}

For all fine-tuning experiments, we used the general HierBERT architecture~\citep{chalkidis2019echrA} with task-specific classification heads on top, as shown in  Figure~\ref{fig:ftarch}.
We experiment with different versions of this architecture which use different PLMs (e.g., LegalBERT, CaseLawBERT, \inlegalbert{}, \incaselawbert{}, \custombert{}, etc.) for the lower-level encoder. The rest of the architecture is kept the same in all versions, for a fair comparison among the different PLMs.

Since the training examples for all end tasks consist of long documents (much beyond the sequence limit of 512 tokens that BERT can accomodate), we divide each document into sentences/chunks~\footnote{Some datasets contain documents already divided into sentences, for others we perform chunking}, and represent the document as a sequence of sentences/chunks.
\[D = \{C_1, C_2, \ldots, C_{|D|}\}, \ C_i = \{t_{i1}, t_{i2}, \ldots, t_{i|C_i|}\}, \ t_{i1} = \texttt{[CLS]}\]
HierBERT consists of a lower-level BERT encoder, which is used to encode individual sentences/chunks.
\[\mathbf h_{i1}, \mathbf h_{i2}, \ldots, \mathbf h_{i|C_i|} \leftarrow BERT(\{t_{i1}, t_{i2}, \ldots, t_{i|C_i|}\})\]
with the \texttt{[CLS]} embedding ($\mathbf h_{i1}$) representing the entire span.
The sequence of such \texttt{[CLS]} embeddings across sentences/chunks are then fed to a higher-level encoder, which in our case is an LSTM module with an Attention Head.
The LSTM generates a sequence of context-aware embeddings (respective to each sentence/chunk).
\[\mathbf h'_1, \mathbf h'_2, \ldots, \mathbf h'_{|D|} \leftarrow LSTM(\{\mathbf h_{11}, \mathbf h_{21}, \ldots, \mathbf h_{|D|1}\})\]
The Attn module produces a single embedding for the entire document from these context-aware sentence/chunk embeddings.
\[\mathbf h'_D \leftarrow Attn(\{\mathbf h'_1, \mathbf h'_2, \ldots, \mathbf h'_{|D|}\})\]
For some tasks, we require a single document-level representation of the entire input, while for others, we require representations for every sentence/chunk.

For every experiment, we also compare  results obtained using our models with the best results reported in the original papers that introduced the datasets we are working with. The Wilcoxon Signed Rank Test has been shown to be effective for most NLP setups~\citep{dror2018sigtest}; we employ it to verify statistical significance of the best results.

\subsection{Legal Statute Identification (LSI)} \label{sec:ft/lsi}

In countries that follow some notion of the Civil Law System, there are written laws (statutes) that guide jurisdiction. 
In court cases of such systems, one or more of these laws are cited, based on their relevance to the facts (evidences) of the case. 
The task of Legal Statute Identification (LSI) aims to automatically identify the relevant statutes (i.e., the statutes that may have been violated) given the facts of the case~\cite{paul2022lesicin,medvedeva2020mlechr,chalkidis2019echrB}.
LSI is widely studied, and is most commonly modeled as a multi-label text classification task.

\subsubsection{Datasets}


We use two different datasets for this task:

\noindent (i)~{\bf \texttt{ILSI}:} 
The Indian Legal Statute Identification (\texttt{ILSI}) Dataset is based on criminal case documents from India~\citep{paul2022lesicin}\footnote{\url{https://github.com/Law-AI/LeSICiN}}. 
The dataset consists of ~65K examples (derived from criminal court cases from the Supreme Court and 6 High Courts of India), split into train, dev and test splits, and a target set of 100 statutes (classes/labels) from the Indian Penal Code. 
All 100 statutes (class labels) are part of the Indian Penal Code, which contains the defined laws for most criminal procedures in India. 
More details of the dataset are available in~\cite{paul2022lesicin}.

\noindent (ii)~{\bf \texttt{ECtHR-B}:} This is the dataset used in the ECtHR Task B from the LexGlue benchmark suite~\citep{chalkidis2022lexglue}\footnote{\url{https://huggingface.co/datasets/lex_glue}}.
The dataset consists of 11K examples (facts from cases argued in the European Court of Human Rights), split into train, dev and test.
The label set consists of 10 articles (classes/labels) from the European Convention of Human Rights, which mainly contain provisions on human rights issues.
Each fact is mapped with the allegedly violated articles from this label set.

\subsubsection{Implementation}

We used the vanilla HierBERT architecture  with a feed-forward network (sigmoid activation) over the document embeddings 
for the final classification.
Both \texttt{ILSI} and \texttt{ECtHR-B} datasets provide the facts segmented into natural sentences. We truncated each sentence (after tokenization) to a maximum of 128 tokens, and chose the first 128 sentences for each document, and these sentences are fed to HierBERT.

All intermediate embeddings have dimensionality of 768.
We experimented with multiple batch sizes \{32, 64, 128\} and found a batch size of 64 to be the most suitable \footnote{By most suitable, we refer to the hyper-parameter setting that produces maximum performance on the respective validation set}.
The model was trained for 25 epochs (with early stopping) with Weighted Binary Cross Entropy Loss, with the label weights being inversely proportional to the label frequency.
We also experimented with a uniform initial learning rate across all layers of 5e-5, as well as different learning rates for different layers (1e-3 for the upper encoder of HierBERT and the classification layer, 1e-5 for the lowest BERT layer) and found the latter more suitable. It took around 40 hrs to run a single experiment for the \texttt{ILSI} dataset and 8 hrs for the \texttt{ECtHR-B} dataset on a single RTX A6000 GPU.

\begin{table*}[!t]
    \centering
    \small
    \begin{tabular}{|l||c|c|c||c|c|c|}
        \hline
        \multirow{2}{*}{\textbf{Encoder Module}} & \multicolumn{3}{c||}{\textbf{\texttt{ISS} dataset}} & \multicolumn{3}{c|}{\textbf{\texttt{UKSS} dataset}} \\ \cline{2-7}
        & \textbf{mP} & \textbf{mR} & \textbf{mF1} & \textbf{mP} & \textbf{mR} & \textbf{mF1} \\\hline
        BERT & 67.56 & 64.21 & 64.41 & 64.28 & 58.51 & 59.54 \\
        LegalBERT & 71.81 & 65.70 & 67.32 & 62.87 & 60.22 & 60.03 \\
        CaseLawBERT & 70.39 & 68.53 & 68.03 & 61.61 & 60.23 & 59.68 \\ \hline
        \inlegalbert{} & 71.65 & 68.36 & 68.98 & 63.93 & \textbf{60.95} & \textbf{61.54} \\
        \incaselawbert{} & 72.20 & 68.21 & 68.80 & 64.68 & 59.82 & 60.85 \\ 
        \custombert{} & 70.9 & 66.91 & 67.25 & 63.28 & 60.41 & 60.62 \\ \hline
        Best result in original paper & \textbf{83.96} & \textbf{80.98} & \textbf{82.08} & \textbf{65.30} & 58.70 & 60.00 \\ \hline
    \end{tabular}
    \caption{Performances for the Semantic Segmentation task. All reported values are macro-averaged and then averaged across 5-folds (cross-validation) and in terms of percentage. Difference in m-F1 results for \inlegalbert{} and all other variants (best fold model) is statistically significant based on the Wilcoxon Signed Rank Test.}
    \label{tab:ssres}
    \vspace{-5mm}
\end{table*}

\subsubsection{Results}

We report Macro-Precision (mP), Macro-Recall (mR) and Macro-F1 (mF1) metrics for comparing the performance of different PLMs (the architectures are identical in all other aspects).
The results over both the datasets are shown in Table~\ref{tab:lsires}.

For the \texttt{ILSI} dataset, the vanilla BERT model, which was pre-trained over only general domain data, performs the poorest across all metrics. 
LegalBERT leverages the pre-trained knowledge of legal data (albeit of a different country), and obtains massive gains over BERT. It also obtains the highest Precision score.
However, CaseLawBERT does even better in terms of F1 over LegalBERT. 
This scenario changes for the \texttt{ECtHR-B} dataset, where we see LegalBERT outperforming CaseLawBERT by some margin. This might be due to the fact that although CaseLawBERT has been trained for more steps on a much larger dataset of US legal documents (helping it to generalize better for the Indian domain), LegalBERT has been trained on ECtHR documents, and thus performs better for the \texttt{ECtHR-B} dataset.

Injecting India-specific legal knowledge further boosts results for \inlegalbert{} across both datasets. 
Although Precision reduces slightly from the vanilla LegalBERT on the \texttt{ILSI} dataset, there are significant improvements for both Recall, and more crucially, F1 score.
There is an outright improvement achieved by \inlegalbert{} across all metrics for the \texttt{ECtHR-B} dataset.
However, \incaselawbert{} shows a slight drop in F1 for both datasets with increase in Recall.
This shows that LegalBERT is more adaptable to further pre-training on data from other countries.
\custombert{} shows comparable performance to both \inlegalbert{} and \incaselawbert{} on \texttt{ILSI} despite being pre-trained for much lesser steps (700k vs. 1.3M vs. 2.3M steps respectively), although \custombert{} uses an Indian-specific vocabulary, which is a better fit for the text of \texttt{ILSI}.
The same vocabulary is not suited for the text of \texttt{ECtHR-B}, and thus the performance for \custombert{} is much lower than \inlegalbert{}.
The difference in results of Table~\ref{tab:lsires} are statistically significant based on the Wilcoxon Signed Rank Test (choice of test based on the observations of \citet{dror2018sigtest}).

\noindent \textbf{Comparison with the original works:} The original work \citep{paul2022lesicin} that introduced the \texttt{ILSI} dataset, employed an architecture (called LeSICiN) which is a hybrid of text-based and graph-based encodings. In particular, they used a Sent2vec encoder~\citep{moghadasi2020sent2vec} (a relatively shallow pre-training approach based on FastText) for encoding text, and reported the best performance of macro-F1 $28\%$.
The models reported in this paper outperform the best result reported in \citep{paul2022lesicin} massively on the \texttt{ILSI} dataset (Table~\ref{tab:lsires}).
We think that transformer-based pre-training, along with some fine-tuning experimental setups, such as different initial learning rates to different kinds of layers, have helped to optimize the models effectively.

On the other hand, \citet{chalkidis2022lexglue} used a very similar architecture to ours for the \texttt{ECtHR-B} dataset, i.e., HierBERT, but with a 2-layer transformer on top instead of LSTM. 
They also reported results using LegalBERT and CaseLawBERT encoders (m-F1 of 74.7\% and 70.3\% respectively), which we beat with our architecture (Table~\ref{tab:lsires}). 
We had also experimented with this version of the HierBERT architecture (with a 2-layer transformer on top) for both datasets, and we observed that the performance was slightly less on \texttt{ECtHR-B} and significantly less on \texttt{ILSI} as compared to the HierBERT version with LSTM + Attn on top.
Thus, we only report the results for HierBERT with LSTM + Attn on top.
Our model with \inlegalbert{} obtains higher performance over all results reported in~\citet{chalkidis2022lexglue} for \texttt{ECtHR-B} -- the macro-F1 obtained by our model is 75.88\% compared to the best value of 74.7\% reported in~\cite{chalkidis2022lexglue}.


\subsection{Semantic Segmentation of case judgements} \label{sec:ft/ss}

Legal case judgements are composed of different functional or \textit{rhetorical segments} such as \emph{Facts}, \emph{Issue}, \emph{Ratio for the ruling}, \emph{Ruling}, etc.~\citep{bhattacharya2019rhetoricalrole}.
Demarcating these segments can be helpful, since many applications require analyzing some specific segments while not considering the rest, e.g., we need to extract only the facts for tasks such as LSI. 
Again, since legal case judgements are very long, law professionals may want to read only certain segments such as only the legal issues in a case.
Legal judgements in many countries do not have clearly demarcated boundaries for the above segments, and thus developing automated methods for semantic/rhetorical segmentation is a very widely researched task. 
The task can be considered as a sequence labeling problem, where the entire document can be considered as a sequence of sentences with each sentence being assigned a particular semantic segment.

\subsubsection{Datasets}
We use two datasets for this task as well:

\noindent (i)~{\bf \texttt{ISS}:} We use the dataset provided by~\citet{bhattacharya2019rhetoricalrole}
and call it the Indian Semantic Segmentation dataset (\texttt{ISS}).\footnote{\url{https://github.com/Law-AI/semantic-segmentation}} 
The \texttt{ISS} dataset has 50 documents from the Supreme Court of India, with each sentence marked with one of seven semantic/rhetorical segment labels -- \textit{Facts}, \textit{Arguments}, \textit{Statutes}, \textit{Precedents}, \textit{Ratio Decidendi}, \textit{Ruling by Lower Court} and \textit{Ruling by Present Court}. 
In total, there are 9,308 sentences marked with one of the labels stated above.

\noindent (ii)~{\bf \texttt{UKSS}:} We also use the dataset based on UK Supreme Court cases provided by~\citet{bhattacharya2021deeprhole}, calling it the UK Semantic Segmentation dataset (\texttt{UKSS}). \footnote{Dataset available upon request to the first author}
The \texttt{UKSS} dataset has 50 documents from the Supreme Court of the UK, segmented into 18,155 sentences, and each sentence is annotated with one of the same seven semantic/rhetorical as the \texttt{ISS} dataset. 

\subsubsection{Implementation}

For this task, we use the HierBERT architecture with Conditional Random Field (CRF) classifier \citep{lafferty2001crf} over the contextualized sentence embeddings, since CRFs are known to work well for sequence labeling tasks.
We limit the sentence length to 128 tokens, but include all sentences in a document.
All intermediate embeddings are of size 768.
We experimented with multiple batch sizes \{1,2,4\} and found a batch size of 2 to be the most suitable.
The model was trained for 25 epochs (with early stopping) with Negative Log Likelihood Loss.
We also experimented with a uniform initial learning rate across all layers of 5e-5, as well as different learning rates for different layers (1e-3 for the upper encoder of HierBERT and the CRF layer, 1e-5 for the lowest BERT layers) and found the latter more suitable.
We used 5-fold cross-validation for both datasets.
It took around 5 hrs to complete all five folds of a single experiment on a single RTX A6000 GPU.

\subsubsection{Results} 

The average results (across 5 folds) for the Semantic Segmentation task on both \texttt{ISS} and \texttt{UKSS} datasets are shown in Table~\ref{tab:ssres}. 
For this task, the trends are very similar for both datasets. 
Taken off-the-shelf, both LegalBERT and CaseLawBERT obtain gains over BERT in terms of F1 score across both datasets, however the difference is larger for \texttt{ISS} than \texttt{UKSS}. 
LegalBERT does better than CaseLawBERT for the UK dataset and vice-versa for the Indian dataset (see Table~\ref{tab:ssres}). 
This observation is similar to what we observed in the LSI task as well (see Section~\ref{sec:ft/lsi}).
On further re-training on Indian data, performance of both LegalBERT and CaseLawBERT increases, however the margins are lesser for CaseLawBERT on both datasets.
\custombert{} slightly underperforms all the other legal PLMs, most possibly due to much lesser no. of steps in pre-training, as discussed in Section~\ref{sec:ft/lsi}.
\inlegalbert{} beats all variants in terms of macro-F1 on both datasets (statistically significant based on Wilcoxon Signed Rank Test).

\noindent \textbf{Comparison with the original works:}
None of the transformer-based models stated above are able to perform as well as the best method in \citep{bhattacharya2019rhetoricalrole} for the \texttt{ISS} dataset, which has the same broad architecture as described above, but uses a Sent2vec encoder~\citep{moghadasi2020sent2vec} (based on FastText) trained on Indian Supreme Court documents instead. 
This Sent2vec-based model has a very high performance across all metrics ($> 80\%$ m-F1) as reported in \citep{bhattacharya2019rhetoricalrole}. 
We suspect this could be due to the very small size of the ISS dataset (which may not be sufficient to fine-tune transformer-based models effectively).
For the \texttt{UKSS} dataset, \citet{bhattacharya2021deeprhole} fine-tuned both BERT and LegalBERT under the same HierBERT-CRF setup as we are using, but had frozen all but the top 2 layers of BERT.
Under our setup (fine-tuning all layers), we achieve similar performance as compared to the result reported by \citet{bhattacharya2021deeprhole} for LegalBERT encoder, which is $60\%$ m-F1 (see Table~\ref{tab:ssres}). 
Using \inlegalbert{} further improves on this performance to achieve $61.54\%$ m-F1 which is higher than the best result reported in~\cite{bhattacharya2021deeprhole}.

        
        
        
        

\subsection{Court Judgment Prediction} \label{sec:ft/cjpe}

Automatically predicting the outcome of court cases has been widely studied in recent years~\citep{zhong-etal-2020-nlp,katz2014usscbehaviour,malik2021ildc}.
This has mostly been done via the task of Legal Judgment Prediction (LJP), where, given the facts of a case, we are to predict the laws/statutes violated, applicable charges and terms of penalty.
Along these lines, the CJPE (Court Judgement Prediction and Explanation) task, introduced by \citet{malik2021ildc}, aims to predict the final decision of the court, based on the rest of the case judgement (i.e., the input is the case judgement document with the final decision removed).
In the Indian scenario, a court case usually consists of one or multiple appeals/claims filed by the appellant against the respondent, and the judge needs to provide an \emph{`accept'} or \emph{`reject'} decision for each of these claims.
Thus, this can be modeled as a binary text classification task.

\subsubsection{Dataset}
The ILDC family of datasets, introduced by \citet{malik2021ildc} contain different types of training data.\footnote{\url{https://github.com/Exploration-Lab/CJPE}}
Among them, the \texttt{ILDC-multi} dataset consists of approx. 35K Supreme Court of India case documents each having multiple claims/appeals; a single label is assigned for each document -- \emph{`accept'} if at least one appeal in that case document is accepted, \emph{`reject'} otherwise.
For our current experiment, we consider the \texttt{ILDC-multi} dataset.



\begin{table}[t]
    \centering
    \small
    \begin{tabular}{|p{0.2\textwidth}|c|c|c|}
        \hline
        \textbf{Encoder Module} & \textbf{mP} & \textbf{mR} & \textbf{mF1} \\\hline
        BERT & 71.31 & 70.98 & 71.14 \\
        LegalBERT & 79.85 & 78.49 & 78.21 \\
        CaseLawBERT & 82.62 & 82.42 & 82.38 \\ \hline
        \inlegalbert & \textbf{83.43} & \textbf{83.15} & \textbf{83.09} \\
        \incaselawbert & 83.05 & 82.82 & 82.77 \\ 
        \custombert & 80.38 & 79.73 & 79.60 \\ \hline
        Best result in original paper & 77.32 & 76.82 & 77.07 \\ \hline
    \end{tabular}
    \caption{Performance over the \texttt{ILDC-multi} dataset for the Court Judgement Prediction task. All values are in terms of percentage. Difference in m-F1 results for \inlegalbert{} and all other variants is statistically significant based on the Wilcoxon Signed Rank Test.}
    \label{tab:cjperes}
    \vspace{-8mm}
\end{table}

\subsubsection{Implementation} \label{sec:ft/cjpe/imp}
We use the HierBERT architecture which was also used by \citet{malik2021ildc} for their results, with a linear layer (sigmoid activation) on top.
The text of the \texttt{ILDC} dataset is not pre-segmented into sentences (unlike all the other datasets described previously), and thus we chunk the entire text (after tokenization) into contiguous, non-overlapping groups of 128 tokens, and passed to the HierBERT model.
We choose the last 128 chunks (based on the observation by \citet{malik2021ildc} that the end portion of documents contains information that can be helpful for CJPE).
All intermediate embeddings are of size 768.
We experimented with multiple batch sizes \{16,32,64\} and found a batch size of 32 to be the most suitable.
The model was trained for 25 epochs (with early stopping) with Binary Cross Entropy Loss.
We also experimented with a uniform initial learning rate across all layers of 5e-5, as well as different learning rates for different layers (1e-3 for the upper encoder of HierBERT, 1e-5 for the lowest BERT layer) and found the latter more suitable. 
It took around 30 hrs to complete a single experiment for the \texttt{ILDC} dataset on a single RTX A6000 GPU.

\subsubsection{Results}

For this task, we again use the vanilla HierBERT architecture with a feed-forward network (sigmoid activation)  over the document embedding, having a single output unit as the final classification layer.
We report macro-P, macro-R and macro-F1 metrics for evaluating the different PLMs in Table~\ref{tab:cjperes}.
Performance of both \legalbert{} and \caselawbert{} improve on re-training over Indian legal data; however, the difference between \legalbert{} and \inlegalbert{} is much greater.
This is consistent with our observations for LSI and Semantic Segmentation that \legalbert{} is more adaptable to re-training on Indian data as compared to \caselawbert{}.
\custombert{} only manages to outperform the vanilla BERT and \legalbert{}, falling significantly shorter than both \inlegalbert{} and \incaselawbert{}.

\noindent {\bf Comparing with the original paper:} 
All 5 legal PLMs outperform the best result reported by \citet{malik2021ildc}, which used the same architecture as ours but employed XLNet \citep{yang2019xlnet} as the lower encoder.
This is possibly due to the legal knowledge accrued by the PLMs we used, given that \citet{malik2021ildc} did not use any legal PLM based encoder.
The m-F1 values achieved by \inlegalbert{} is \textit{statistically significantly} higher than that of all other models and the results in the original papers, based on the Wilcoxon Signed Rank Test.



\subsection{Analysis of Explainabilty of Models}



Explainability of automated methods is extremely crucial in the legal domain, to be accepted as authentic by both legal professionals and laymen.
For domain-specific tasks like Court Judgment Prediction, this is even more important, since the standard output of the model is just a `yes'/`no' decision.
For instance, it is important to understand if the model is basing its predictions over the appropriate portions of the input text.
To this end, we now check the agreement of the various fine-tuned models (that use the different PLMs) with the opinions of domain experts for the CJPE task.

\subsubsection{Dataset}
The \texttt{ILDC} dataset for CJPE~\cite{malik2021ildc} includes a small set of 56 documents (taken from the test set), where \textit{five legal experts have manually annotated the spans of text (character-level spans) which are crucial for understanding the final judgment}.
We make use of this set, called \texttt{ILDC-expert} for understanding the predictions of the model, which contains five different sets of annotations, corresponding to each expert, per document.

\subsubsection{Method}
We now describe how we check the agreement of the models with the annotations by the legal experts.

\noindent {\bf Weightage assigned by fine-tuned models:} As discussed in Section~\ref{sec:ft/cjpe/imp}, we divide the entire text of each document $D$ into equal-sized chunks and feed these in hierarchical fashion to the models, both during training and testing.
To understand the relative weightage placed by a fine-tuned model $\mathcal M$ over different parts of the text for building the final document representation, we can use the attention scores generated by the upper (BiLSTM + Attn) encoder (shown in Figure~\ref{fig:ftarch}).
For a document $D$, the upper encoder of $\mathcal M$ generates a probability distribution $\{p_{D,i}^{\mathcal M}\}$ over all chunks $C_{D,i}^{\mathcal M}$ of $D$ ($i \in [1,n]$, where $n$ is the total number of chunks).
Note that these chunks (of tokens) are not universal --- they depend on the tokenizer used, and thus, depend on the model $\mathcal M$. 

\noindent {\bf Weightage assigned by experts:} For each document $D$, an expert $\mathcal E$ provides one or more contiguous gold-standard character spans for the entire document, marking the parts of the text that guide the final decision of the court.
For comparing these annotations of expert $\mathcal E$ with the attention weightage generated by a model $\mathcal M$, we need to find some notion of chunk-level gold-standard importance based on the annotated spans (i.e., some estimate of which chunks would have been deemed important by experts).

We apply the following procedure to find the gold-standard, chunk-level importance scores of expert $\mathcal E$ with respect to model $\mathcal M$. For every document $D$,

\noindent~(i) We tokenize the entire text of $D$ into a sequence of tokens $\mathcal T_D^{\mathcal M}$  and map the character-level spans to token-level spans , based on the tokenizer of $\mathcal M$.
This effectively gives us a mapping $\mathcal R^{\mathcal E}: \mathcal T_D^{\mathcal M} \rightarrow \{0,1\}$, from each token $t \in \mathcal T_D^{\mathcal M}$ to a binary value of 0 (if $t$ is not part of a span that was annotated by the expert) or 1 (if $t$ is part of an annotated span).

\noindent~(ii) For each chunk $C_{D,i}^{\mathcal M} \in D$, we can count the number of annotated tokens (a token $t$ is annotated iff $\mathcal R^{\mathcal E}(t) = 1$) inside it. 
Dividing this by the total number of annotated tokens (by expert $\mathcal E$) in the whole document $D$, gives us the gold-standard distribution $\{q_{D,i}^{\mathcal M, \mathcal E}\}$ of importance over the chunks, for each chunk $C_{D,i}^{\mathcal M} \in D$. 

Thus, following the above process, we obtain, for each document $D$, two distributions ---
(i)~the model-assigned attention distribution $\{p_{D,i}^{\mathcal M}\}$, and,
(ii)~the expert-assigned importance distribution $\{q_{D,i}^{\mathcal M, \mathcal E}\}$,
for every expert $\mathcal E$ and every model $\mathcal M$.
To measure the agreement of expert $\mathcal E$ with the model $\mathcal M$, we 
compute the KL-Divergence score between these two distributions. Note that a lower value of the KL-divergence indicates a better match between the expert and the model.

\begin{table}[!t]
    \centering
    \small
    \begin{tabular}{|p{0.12\textwidth}|c|c|c|c|c|c|}
        \hline
        \textbf{Model} & \textbf{E1} & \textbf{E2} & \textbf{E3} & \textbf{E4} & \textbf{E5}  & \textbf{Average} \\ \hline
        \legalbert{} & 47.4 & 44.7 & 44.7 & 42.8 & 44.8 & 44.9 \\ 
        \caselawbert{} & 55.2 & 50.5 & 48.1 & 47.1 & 53.6 & 50.9 \\ \hline
        \inlegalbert{} & 44.4 & 43.5 & 43.8 & 41.1 & 45.4 & 43.6 \\
        \incaselawbert{} & \textbf{40.7} & \textbf{42.4} & \textbf{43.1} & \textbf{40.9} & \textbf{40.7} & \textbf{41.6} \\ 
        \custombert{} & 43.7 & 47.3 & 47.0 & 43.9 & 40.5 & 44.5 \\ \hline
    \end{tabular}
    \caption{KL Divergence of the model-assigned chunk attentions for \legalbert{}, \inlegalbert{} and \custombert{} w.r.t the probabilities of annotated tokens in every chunk. Columns represent each of the 5 legal experts who have provided the annotations.}
    \label{tab:kldiv}
    \vspace{-10mm}
\end{table}

\subsubsection{Results}
We average the KL-divergence scores across all $D \in \texttt{ILDC-expert}$ (set of 56 documents) to obtain a single score for a given expert $\mathcal E$ and a given model $\mathcal M$.
This process is repeated across all experts and all models and the KL-divergence values are stated in Table~\ref{tab:kldiv}.
For 4 out of 5 experts, we see that \inlegalbert{} has a lower KL-divergence score than \legalbert{}, whereas \incaselawbert{} has a lower score than \caselawbert{} for all 5 experts.
This corroborates with the fact that CJPE performance (macro-F1 over the test set of \texttt{ILDC-multi}, reported in Table~\ref{tab:cjperes}) improves upon re-training on Indian data.
Both \inlegalbert{} and \incaselawbert{} achieve lower scores than \custombert{} for majority of the experts.

Interestingly, \caselawbert{} has comparatively poor scores of KL-divergence despite having a decent performance in terms of macro-F1.
Also, \inlegalbert{} attains a better macro-F1 score than \incaselawbert{} (Table~\ref{tab:cjperes}), but \incaselawbert{} shows greater agreement with the experts than \inlegalbert{} (Table~\ref{tab:kldiv}).
It is noteworthy that the macro-F1 has been calculated over the test set in the \texttt{ILDC-multi} dataset, whereas the KL-Divergence scores are calculated over \texttt{ILDC-expert}, which is a small subset of the former.


\section{Conclusion and Future Work} \label{sec:conc}
In this work, we have investigated the gains of re-training existing pre-trained models for the legal domain, namely, LegalBERT~\citep{chalkidis2020legalbert} and CaseLawBERT~\citep{zheng2021legalbertus}, on a large corpus of Indian legal data.
We also pre-trained a BERT-base model from scratch on the same corpus.
Based on the results across three practically important legal NLP tasks over standard datasets from India as well as from other domains/countries --- Legal Statute Identification, Semantic Segmentation, and Court Judgement Prediction -- we can conclude that domain-specific pre-training on Indian legal text helps, in general, to improve the performances for both the above models.

We observe some common trends about various PLMs in the legal domain:
(i)~When re-trained on Indian data, \inlegalbert{} improves significantly over LegalBERT, while the gains are much smaller for \incaselawbert{} over CaseLawBERT for Indian datasets. 
In fact, \inlegalbert{} attains the best F1 scores across all the three tasks and five datasets, which shows that the LegalBERT model is more adaptable to re-training on new data, as compared to the CaseLawBERT model, while also improving in performance on old (EU) data.
(ii)~\custombert{} comes close in terms of macro-F1 to both \inlegalbert{} and \incaselawbert{} across multiple datasets despite having being trained for much lesser number of steps (700k vs. 1.3M vs. 2.3M steps respectively).
(iii)~For 4 out of the 5 datasets we have experimented with, \inlegalbert{} (available at \url{https://huggingface.co/law-ai/InLegalBERT}) outperforms the result in the original papers (which introduced the respective dataset).

In general, these experiments show that there is promise in developing country-specific legal PLMs, especially if there are linguistic idiosyncracies (as seen in Indian legal documents).
So, what should be the prospective way of developing such a PLM --
(i) Re-training an existing legal PLM on the domain-specific data, or
(ii) Pre-train an LM from scratch, preferably using domain-specific vocabulary?
Although prior works have shown that the latter is usually a better approach~\citep{chalkidis2020legalbert,zheng2021legalbertus}, the choice might actually be dependent on factors such as size of pre-training data, or the availability of computing resources.
In case the number of pre-training steps is limited, re-training an existing legal PLM might be a promising approach, to obtain improvements over the vanilla variants.
Since training from scratch naturally requires a lot of pre-training steps, this approach can underperform in the case of a limited pre-training budget.
Moreover, re-trained PLMs have the advantage of having being trained on legal data from multiple countries.


The PLMs \inlegalbert{}, \incaselawbert{} and \custombert{}, as well as the code for pre-training and fine-tuning are available publicly to the research community (see Section~\ref{sec:intro}).


In the future, we plan to further investigate with more sophisticated architectures for pre-training.
In particular, hierarchical encoding methods can be useful for exploiting the structure of the text of legal documents.
We also wish to conduct more experiments to investigate the pre-existing biases in the PLMs, and how these are affected by re-training on Indian data.

\vspace{2mm}
\noindent {\bf Acknowledgements:}
Shounak Paul is supported by the Prime Minister's Research Fellowship (PMRF) from the Government of India. 
We acknowledge National Supercomputing Mission (NSM) for providing computing resources of `PARAM Shakti' at IIT Kharagpur, which is implemented by C-DAC and supported by MeitY and DST, Government of India.



\bibliographystyle{ACM-Reference-Format}
\bibliography{sample-refs}



\end{document}